# Strawberry detection and counting based on YOLOv7 pruning and information based tracking algorithm

Shiyu Liu[a, 1], Congliang Zhou[b, 1, *], Won Suk Lee[b]

[a]Department of Electrical and Computer Engineering, Gainesville, Florida, USA

[b]Department of Agricultural and Biological Engineering, University of Florida, Gainesville, Florida, USA

[1] Equal contributions.

* Corresponding author.

E-mail address: co.zhou@ufl.edu (C. Zhou).

## Abstract

The strawberry industry yields significant economic benefits for Florida, yet the process of monitoring strawberry growth and yield is labor-intensive and costly. The development of machine learning-based detection and tracking methodologies has been used for helping automated monitoring and prediction of strawberry yield, still, enhancement has been limited as previous studies only applied the deep learning method for flower and fruit detection, which did not consider the unique characteristics of image datasets collected by the machine vision system. This study proposed an optimal pruning of detection heads of the deep learning model (YOLOv7 and its variants) that could achieve fast and precise strawberry flower, immature fruit, and mature fruit detection. Thereafter, an enhanced object tracking algorithm, which is called the Information Based Tracking Algorithm (IBTA) utilized the best detection result, removed the Kalman Filter, and integrated moving direction, velocity, and spatial information to improve the precision in strawberry flower and fruit tracking. The proposed pruning of detection heads across YOLOv7 variants, notably Pruning-YOLOv7-tiny with detection head 3 and Pruning-YOLOv7-tiny with heads 2 and 3 achieved the best inference speed (163.9 frames per second) and detection accuracy (89.1%), respectively. On the other hand, the effect of IBTA was proved by comparing it with the centroid tracking algorithm (CTA), the Multiple Object Tracking Accuracy (MOTA) and Multiple Object Tracking Precision (MOTP) of IBTA were 12.3% and 6.0% higher than that of CTA, accordingly. In addition, other object-tracking evaluation metrics, including IDF1, IDR, IDP, MT, and IDs, show that IBTA performed better than CTA in strawberry flower and fruit tracking. Overall, this study not only proposed a model with higher speed and accuracy but also demonstrated that utilizing the characteristics of the agriculture dataset can achieve more efficient object detection and tracking.

## 1. Introduction

Strawberry yield mapping and prediction can help growers hire an optimal number of people for harvesting and make optimal marketing decisions. However, manually counting fruits and flowers for yield mapping and prediction proves to be cost-ineffective. In the United States, Florida is confronted with increasing labor costs, averaging between $8,000 and $10,000 per acre (Guan et al 2018). Labor cost is around 40% of total Florida strawberry production costs, which makes Florida hard to compete with other countries with cheap labor, such as Mexico and China (Suh et al 2017).

Recently, with the development of machine learning and computer vision, automatically detecting and counting fruits and flowers has become possible and shows the potential to reduce the labor requirement for yield mapping and prediction in the strawberry field. The Viola-jones framework (Viola & Jones 2001) marked a breakthrough in enabling real-time detection of human faces without any constraints. Since 2013, OverFeat (Sermanet et al 2013) has been proposed as a pioneering algorithm, followed by various object detection algorithms, including the R-CNN series (Girshick et al 2014; Girshick 2015; Ren et al 2015) and YOLO series (Redmon et al 2016; Redmon & Farhadi 2017; Redmon and Farhadi 2018; Bochkovskiy et al 2020; Li et al 2022; Wang et al 2023). These object detection algorithms show great potential in solving agricultural problems such as crop detection (Puranik et al 2021), crop count estimation (Immaneni & Chang 2022), pest control (Zhou et al 2023a), disease detection (Xiao et al 2020), and smart farming (Attri et al 2023). Meanwhile, many researchers modified the architecture of deep learning models to achieve enhanced speed and increased precision in agricultural applications. For example, Jiang et al (2022) modified YOLOv7 by incorporating an attention mechanism, aimed at improving the accuracy of hemp duck count estimation. Zhang et al (2022) modified the backbone of YOLOv4 to achieve strawberry detection in real-time.

However, previous studies mainly focused on improving feature extraction and fusion for object detection but ignored that deep learning-based object detection methods usually use multiple detection heads to detect objects at different scales. Since YOLOv3, YOLO has developed three to four detection heads to generate various sizes of output to fit different scale object detection. Other object detection methods, such as faster-RCNN (Ren et al 2015), output four scales feature maps to improve the detection accuracy. Detection heads with proper scales can fully extract the feature information to increase the accuracy of detection to a certain extent (Yu et al 2023). Besides, agricultural datasets collected by machine vision systems have their unique characteristics. For example, the machine vision system mounted the camera at a specific height to collect strawberry images, resulting in the size of strawberry flowers and fruits in these images exhibiting minimal variation ([Lin et al 2020](); [Yu et al 2019; Zhou et al 2023b; Chen et al 2019]()). This indicated that employing multiple detection heads for strawberry flower and fruit detection is superfluous; instead, reducing the number of detection heads will lead to structure simplified, and detection speed enhanced.

Following object detection, object tracking methods monitor the movement of the object in a sequence of frames or images, which find applications in motion-based recognition ([BenAbdelkader et al 2002]()), automated surveillance ([Javed & Shah 2002]()), traffic monitoring ([Ferrier et al 1994]()), and vehicle navigation ([Leonard & Bahr 2016]()). The representative baseline methods include simple online and real-time tracking (SORT) ([Bewley et al 2016]()), Deep SORT ([Wojke et al 2017]()) Centroid tracking algorithm ([Puranik et al 2021]()), and so on. These days, object tracking has been widely utilized in agriculture for yield estimation, exemplifying its practical use in counting strawberry fruits and flowers. For instance, [Heylen et al 2021]() trained a deep-learning model to count strawberry flowers on drone imagery and the model achieved a similar accuracy as manual counting. Traditional object-tracking methods require a significant overlap of the same object between two successive images and the moving direction of objects is not constant ([Li et al 2015]()). In contrast, the machine vision system in this study was moving in a specific direction and at a constant speed but with low frame rates, leading to non-overlapping images of the flowers or fruits in successive frames. However, the moving direction of objects was known.

This study aimed to develop a simple and accurate object detection and tracking method for strawberry yield mapping and prediction. The three objectives of this study were (1) to compare different deep learning models (YOLOv7, YOLOv7-tiny, and YOLOv7-X) for flower and fruit detection, (2) to identify the optimal number and combination of detection heads of deep learning model based on the accuracy and speed of strawberry fruit and flower detection, and (3) to improve the tracking accuracy of strawberry flower and fruit by incorporating spatial position information of objects and known moving speed and direction information of our machine vision system.

## 2. Materials and Methods

### 2.1. Image acquisition and labeling

Two different datasets were collected in this study by using an autonomous robot equipped with four RGB cameras, moving at a constant speed for strawberry flower and fruit image acquisition. Detailed information about the imaging robot can be found in the previous research (Zhou et al 2023b). The first dataset included a total of 2291 images for training and testing deep-learning models for strawberry flower and fruit detection (Fig. 1), while the second dataset consisted of 55 videos, each video included 100 frames of strawberry plants for object tracking testing.

The two datasets were labeled for model training and testing. Labels for object detection were generated manually by LableImg, which is an open-source tool on GitHub. Following the classification criteria proposed by Zhou et al 2021, the objects were categorized into flowers, immature fruits, and mature fruits. With the assistance of LabelImg, .xml files that contained the location information and object labels were generated. The .xml files were converted into .txt files, where"0", "1", and "2" corresponded to "flower", "immature fruit", and "mature fruit," respectively. The labeled image datasets were partitioned into three distinct subsets, namely, training, validation, and test datasets, employing a proportionate allocation ratio of 8:1:1. Mosaic data augmentation was used to augment the dataset. The workflow of object detection data pre-processing is shown in Fig. 1. Labels for object tracking were generated manually by an open-source tool on GitHub called Darklabel. The objects were labeled separately, and the label number started from 1. As a result, three .txt documents were generated for each video, which stood for flower, immature fruit, and mature fruit, respectively. The detail of the tracking label files is shown in Fig. 2.

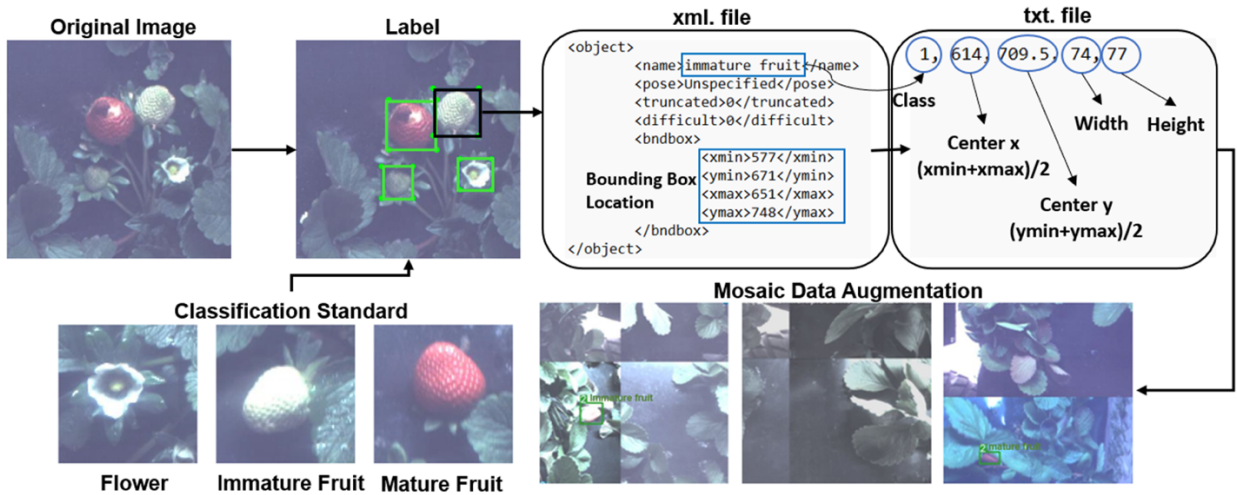

Fig. 1. Workflow of object detection data pre-processing.

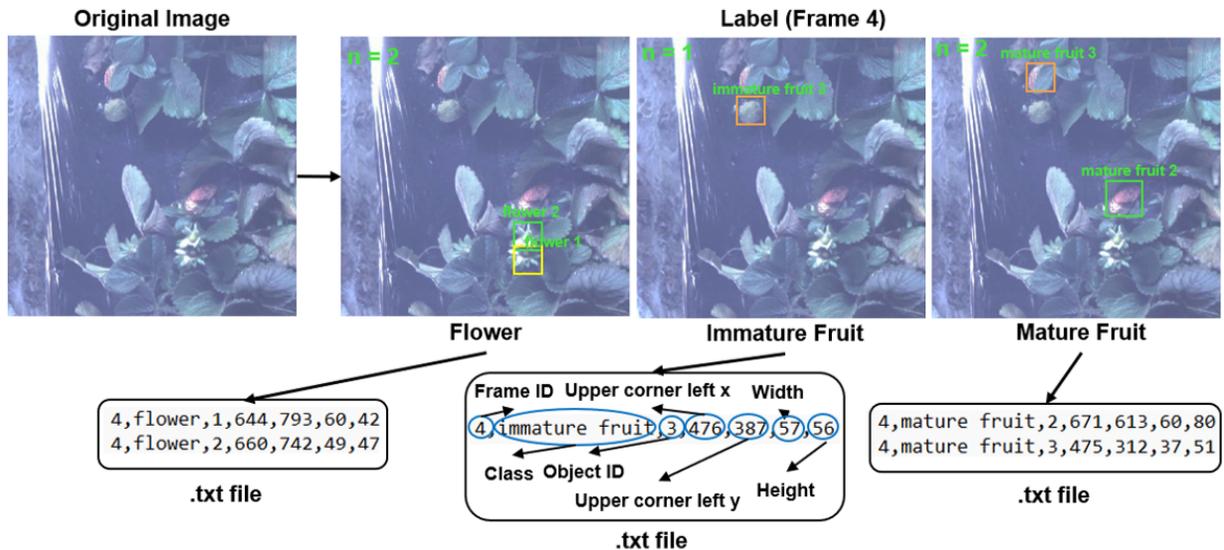

Fig. 2. Workflow of object tracking data pre-processing.

## 2.2. Object detection model

### 2.2.1. YOLOv7 model

The YOLOv7 series of object detection models developed by Wang, et al 2023 has outperformed other widely used object detectors. To meet the needs of various application scenarios, YOLOv7-tiny and YOLOv7-W6 base models were designed for edge GPU and cloud CPU, respectively. Notably, YOLOv7-W6 incorporates auxiliary heads as deep supervision, which is a training methodology that involves introducing auxiliary loss functions at multiple intermediate layers of the network to guide feature learning and improve model performance. YOLOv7-tiny employs leaky ReLU as an activation function, while SiLU is used as an activation function for other models. YOLOv7-X is obtained by combining YOLOv7 with a compound scaling method. YOLOv7-E6 and YOLOv7-D6 are designed by combining YOLOv7-W6 with the same scaling method. YOLOv7-E6E is designed by integrating extended efficient layer aggregation networks (E-ELAN) into YOLOv7-E6, where E-ELAN is proposed to enhance the learning ability of the model.

After analyzing the network structure, YOLOv7-W6, YOLOv7-D6, YOLOv7-E6, and YOLOv7-E6E were all found to have four lead detection heads corresponding to four different scales. Meanwhile, they also have four auxiliary detection heads to assist in label assignment during the training phase. Given the simplicity of the strawberry dataset which features only three classes and minimal size variation among objects, the excessive number of detection heads in these models introduces a high level of complexity, contradicting the objective of training a simple and efficient model for strawberry flower and fruit detection. Therefore, in the follow-up experiments, YOLOv7, YOLOv7-tiny, and YOLOv7-X models were selected as the baselines, which have three distinct scale detection heads and do not include any auxiliary detection heads.

The detailed parameters of YOLOv7 and its variants are shown in [Table 1]. To compare the complexity between models, the floating-point operations (FLOPs) were calculated, which represented the computational workload of the arithmetic operations of the model.

*Table 1 Parameters of YOLOv7 and its variants*

| Model | E-ELAN | Activation function | Compound scale | Auxiliary head | Baseline | Anchor number |
|-------|--------|---------------------|----------------|----------------|----------|---------------|
| YOLOv7-tiny | - | Leaky ReLU | - | - | - | 3 |
| YOLOv7-W6 | - | SiLU | - | Yes | - | 4 |
| YOLOv7 | - | SiLU | - | - | - | 3 |
| YOLOv7-E6 | - | SiLU | Yes | Yes | YOLOv7-w6 | 4 |
| YOLOv7-D6 | - | SiLU | Yes | Yes | YOLOv7-w6 | 4 |
| YOLOv7-X | - | SiLU | Yes | - | YOLOv7 | 3 |
| YOLOv7-E6E | Yes | SiLU | Yes | Yes | YOLOv7-E6 | 4 |

## 2.2.2. Model pruning

The minimal size variation among strawberry flowers and fruits on the RGB image made it feasible to reduce the number of detection heads without compromising detection accuracy. YOLOv7-tiny, YOLOv7, and YOLOv7-X were selected as the baseline models and their structure were modified to encompass all possible combinations of detection heads. Each original model was equipped with three detection heads, resulting in seven combinations. For ease of reference, each detection head was assigned a serial number. According to the Microsoft COCO dataset (Lin et al., 2015), objects can be classified into small objects, medium objects, and large objects. The detection head that focused on detecting larger objects, associating with the large anchor, and generating a small feature map, was denoted as "head 1". Similarly, the detection head responsible for detecting medium-sized objects was referred to as "head 2", and the final head tasks with detecting smaller objects were designated as "head 3". The reduction in detection heads led to a simplification of the model structure, resulting in more efficient models.

Taking YOLOv7-tiny as an example, Fig. 3 shows the detailed structure of the original YOLOv7-tiny, and Fig. 4 shows Pruning-YOLOv7-tiny with different detection head combinations. The model structure of the original YOLOv7, YOLOv7-X, Pruning-YOLOv7, and Pruning-YOLOv7-X can be found in the appendices. To comprehensively evaluate the efficacy of YOLOv7 and its derivative architectures on the strawberry dataset, this research trained the models based on the pre-trained models with the strawberry data, followed by selecting the optimal hyperparameters for performance assessment during the testing phase. The object detection experiments were based on the environment shown in Table 2

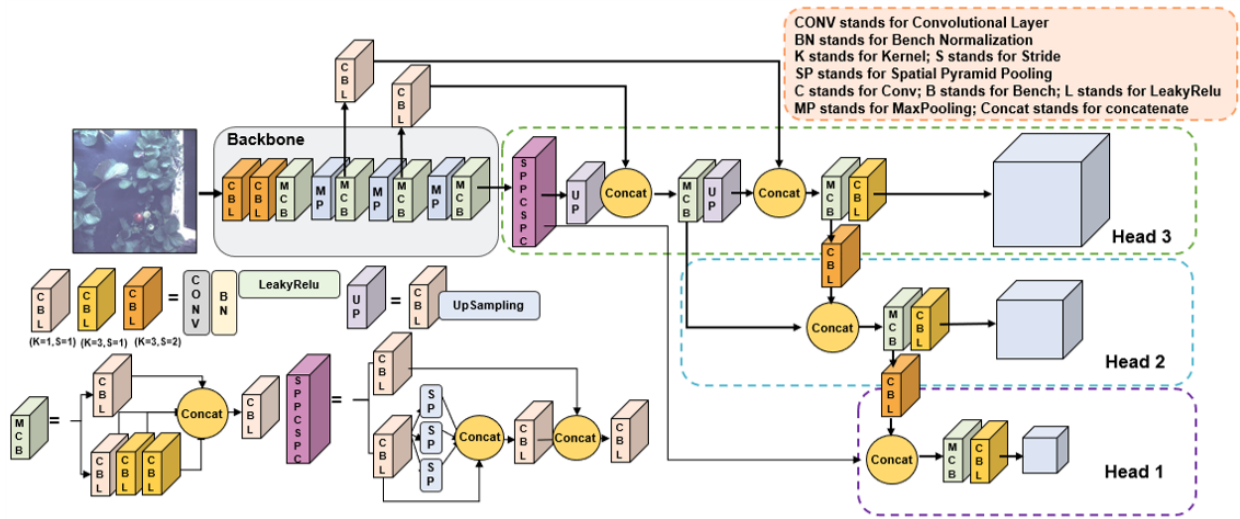

Fig. 3. YOLOv7-tiny structure.

*Table 2 Experiment configuration for object detection*

| Component | Parameter |
| --- | --- |
| Operating system | Ubuntu 22.0 |
| Operating system type | 64-bit |
| GPU | GeForce GTX Titan X |
| Accelerated environment | CUDA 11.5 |
| Framework | PyTorch 1.11.0 |
| Integrated Development Environment (IDE) | PyCharm 2022.3 |

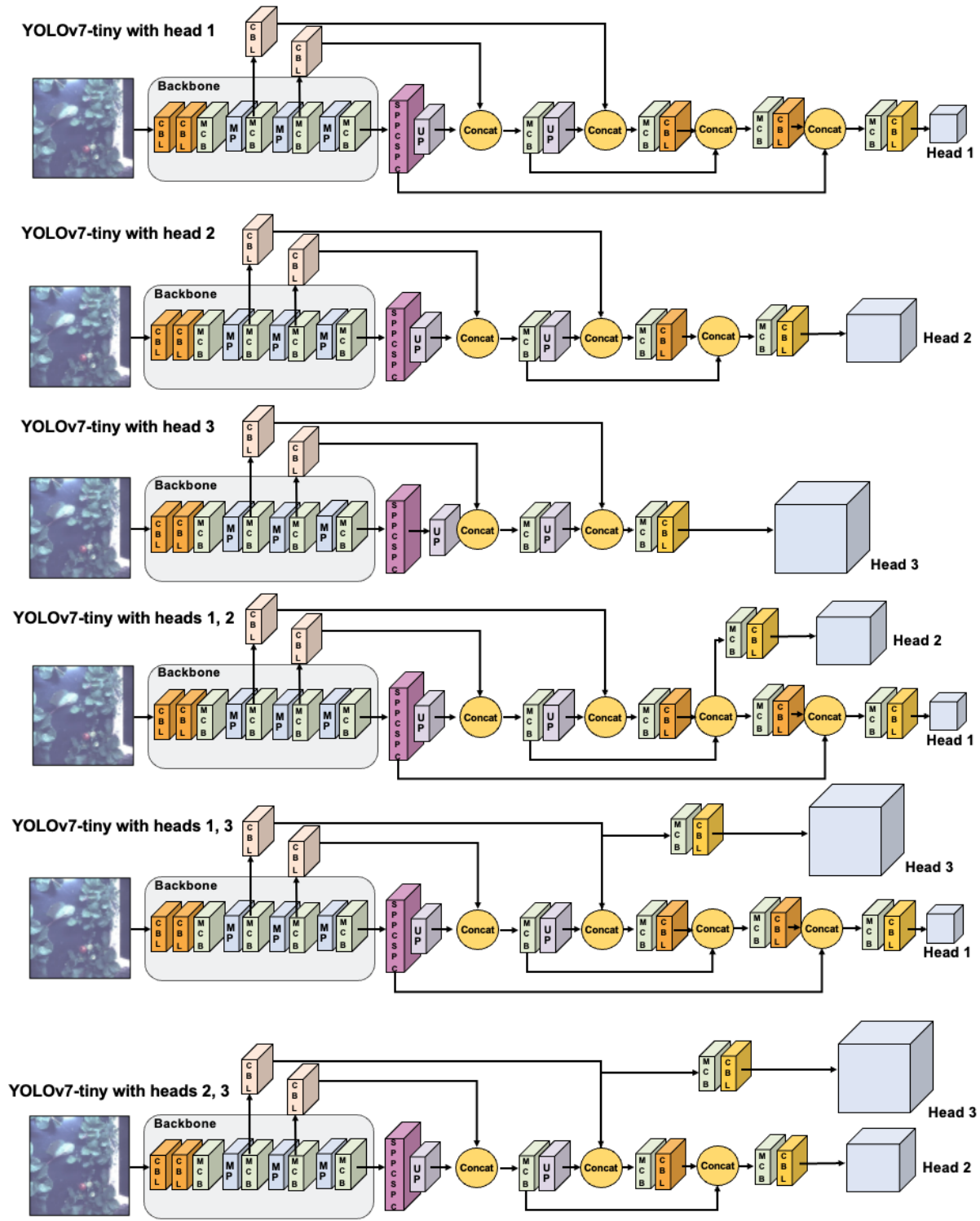

Fig. 4. Pruning-YOLOv7-tiny with separate head structure.

## 2.3. Information Based Tracking Algorithm (IBTA)

### 2.3.1. Baseline: Simple Online and Realtime Tracking (SORT)

IBTA was based on the SORT algorithm and made use of all the known information about the strawberry dataset, including the speed and direction of objects moving, and the spatial position of objects within the same frame to make a more accurate tracking of the objects. The workflow of SORT and IBTA is shown in Fig. 5. Compared with SORT, IBTA removed the Kalman Filter to simplify the process, as it predicted the position of the flower and fruit in the next frame with the speed and direction information of the robot, and the spatial information between objects in one frame. After getting the detection result from YOLOv7 and its variants, a location score was calculated to represent the spatial relationship between the objects in one frame. The preset robot's velocity and direction were utilized to predict the location of the flower and fruit in the next frame. During the matching, the IOU value and the location score were both used. The IOU between the objects was calculated using Eq. (1).

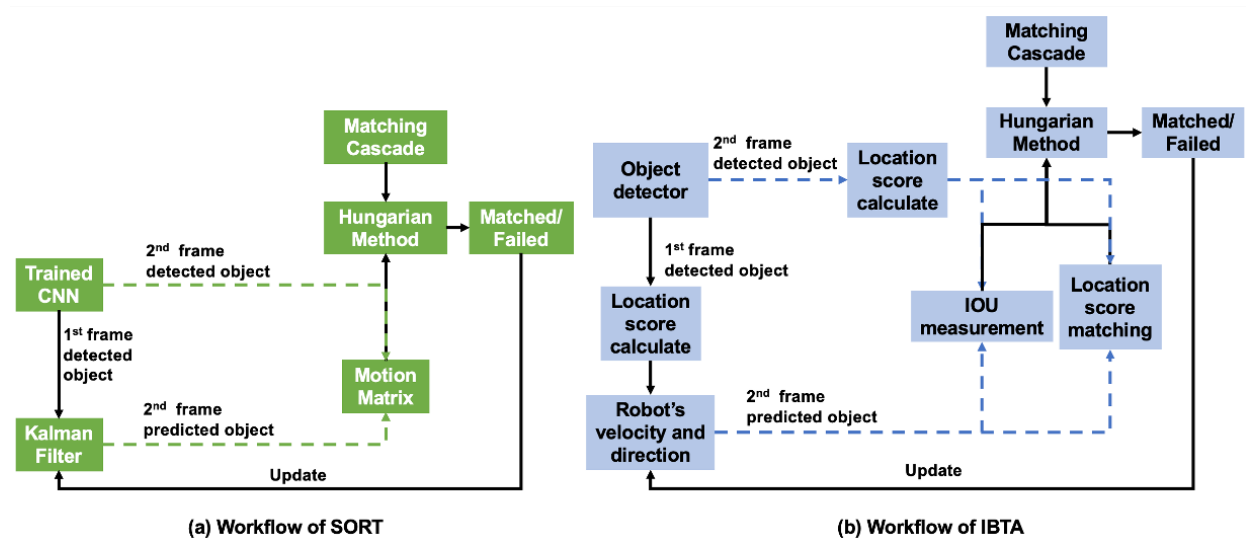

Fig. 5. Flowchart of SORT and IBTA.

$$IOU = \frac{A \cap B}{A \cup B} \qquad (1)$$

Where A and B are bounding boxes of two objects.

## 2.3.2. Location score: spatial relationship between objects

The spatial position of objects within the same frame can be utilized to avoid re-counting and to address the overlap problem. It is common for two fruits to be close together in strawberries, leading to the situation where, as the robot moves, the smaller fruit might move to have a larger overlap with the larger ones. In an extreme case, if a smaller object was completely hidden by a larger one and went undetected, it can lead to a recount when it reappears. Therefore, with the knowledge that there were two objects close together in a frame, it was expected that these two objects would appear together in the following frame. In a case where the second frame experiences detection failure or one of the objects is obscured, it can be inferred based on the spatial position information of objects from the previous frame. The utilization of spatial position information was presented by calculating the location score for each object in the image. If the IOU of two bounding boxes was larger than the threshold, the size of the two objects would be compared, and the larger one would be set as a standard. Subsequently, the small object that appeared on the large one's left, central, and right would be assigned a location score of "1", "2", and "3", respectively. "0" was assigned to the larger object as its location score, and the objects that didn't have an adjacent neighbor were assigned "None". If multiple objects were clustered together, location scores for the largest and second-largest objects were first calculated and then performed the matching process. After matching these objects, the remaining ones were sorted by size and assigned location scores. The object moving in the strawberry dataset and the location assignment examples are shown in Fig. 6.

Considering the noise caused by the uneven ground, wind speed, and other possible uncertainty, a buffer was set up to expand the range of predicted object locations. In actual operation, the raw location in the next frame was calculated by adding the speed of the robot to the y-axis value of the object's current location and keeping the x-axis value. In Fig.6, the buffer was the yellow dotted line circular area, in which the center was the raw predicted location, and the radius r was the maximum moving speed minus the average moving speed.

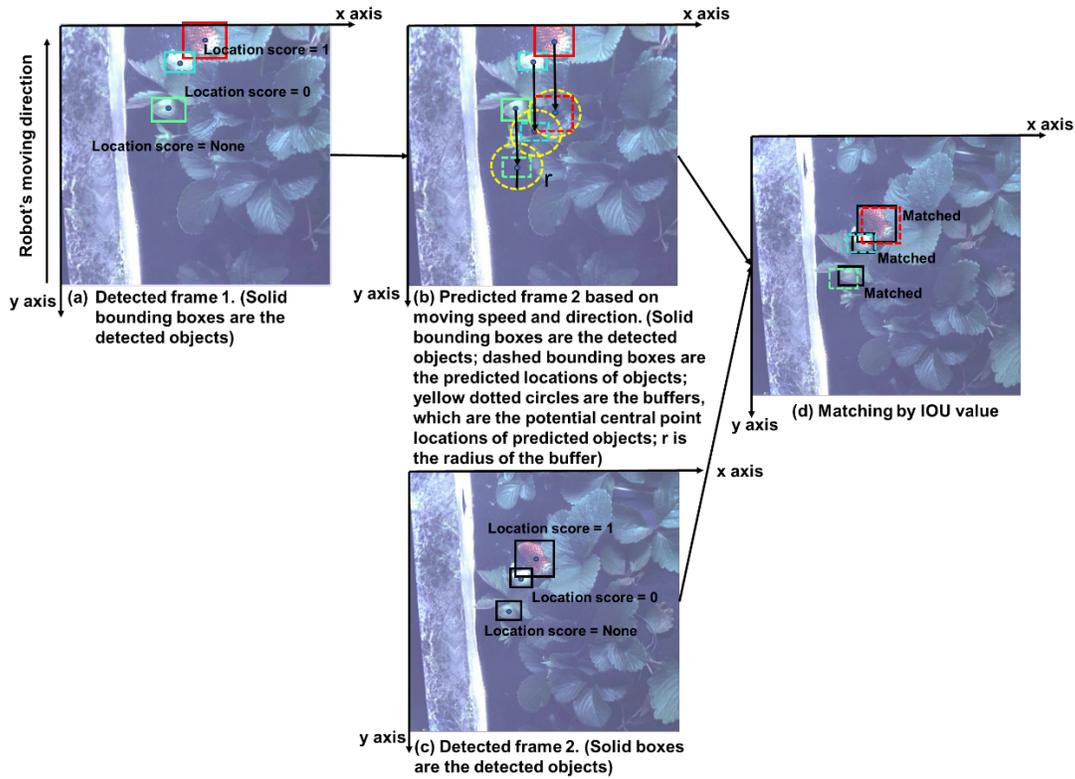

Fig. 6. Image Demonstration of IBTA.

### 2.3.3. Matching cascade

Two lists were utilized to track objects, including a detection list and a tracking list. The detection list contained objects detected in the current frame, while the tracking list comprised objects that were being continuously monitored. The matching process between these lists was governed by assigning location scores to each object and comparing the IOU between the two lists to a threshold. Objects from the detection list with IOU values surpassing a predefined IOU threshold were considered matched and consequently removed from the list, paving the way for the next object to undergo matching. Unmatched objects were treated differently based on their size: larger objects that did not match can be either new observations or ones that were missed in earlier frames. In contrast, for smaller objects that did not match, the treatment depended on whether their larger paired object was matched or not. If a larger object was matched, its smaller counterpart was also classified as matched, presuming potential occlusion or prior missed detection. Conversely, if a larger object remained unmatched, the smaller object was similarly categorized as unmatched, indicating a probable new detection. The object tracking experiments were based on the environment shown in Table 3.

*Table 3 Experiment configuration for object tracking*

| Component | Parameter |
| --- | --- |
| Operating system | Windows 11 |
| Operating system type | 64-bit |
| GPU | GeForce GTX 1650 |
| Interpreter | Python 3.6 |
| Integrated Development Environment (IDE) | PyCharm 2011.3.2 |

## 2.4. Evaluation methods

To evaluate the object detection model, mean average precision ($mAP_{0.5}$) (Padilla et al 2020) and frame per second (FPS) were used in this study. $mAP_{0.5}$ is a measure of accuracy that averages the Average Precision (AP) across all classes, considering only detections where the predicted bounding box overlaps at least 50% with the ground truth, where AP is the area beneath the Precision-Recall curve (Sofaer et al 2019). FPS was calculated on the test dataset, which indicates the inference speed of the model. The calculation of FPS is presented by Eq. (2), where inference time does not include the pre-processing and post-processing time.

$$FPS = \frac{1}{inference\ time(s)} \tag{2}$$

Object tracking evaluation metric includes ID Precision (IDP) (Eq. (3)), ID Recall (IDR) (Eq. (4)), IDF1 (Eq. (5)), Mostly Tracked (MT), ID Switches (IDs), Multiple Object Tracking Accuracy (MOTA) (Eq. (6)), and Multiple Object Tracking Precision (MOTP) (Eq. (7)).

$$IDP = \frac{IDTP}{IDTP + IDFP} \tag{3}$$

$$IDR = \frac{IDTP}{IDTP + IDFN} \tag{4}$$

$$IDF1 = \frac{2}{\frac{1}{IDP} + \frac{1}{IDR}} \tag{5}$$

$$MOTA = 1 - \frac{IDFN + IDFP + IDs}{GT} \tag{6}$$

$$MOTP = \frac{\sum_{t,i} d_{t,i}}{\sum_t c_t} \tag{7}$$

Where IDTP represents the correctly tracked targets; IDFN characterizes the targeted objects that are missed; IDFP presents instances where objects that are not for tracking are erroneously subjected to tracking processes; GT corresponds to ground truth. $d_{t,i}$ presents the overlap between the actual and predicted bounding boxes for the $i^{th}$ object at time $t$. $c_t$ are the instances where the tracking algorithm correctly matched a target across time $t$.

## 3.   Result and Discussion

### 3.1. Data Analysis

To explore the relationship between the object's size, the detection head's scale, and the detection performance, the number and size of the objects in the image dataset were analyzed. The size classification results of the objects in the strawberry data set are shown in Table 4. The size distribution of all object types concentrated on small and medium-sized objects, with no large objects present, which suggested that a reduction in the number of detection heads of YOLOv7 and its variants might not necessarily compromise its detection accuracy. Moreover, there is a probability that such a reduction could potentially enhance performance.

*Table 4 The number of target objects*

| Object | Small | Medium | Large |
|--------|-------|--------|-------|
| Flower | 575 | 2198 | 0 |
| Immature fruit | 1213 | 1430 | 0 |
| Mature fruit | 75 | 635 | 0 |

### 3.2.   Deep learning models (YOLOv7, YOLOv7-tiny, and YOLOv7-X) comparison

Table 5 shows that the value of FLOPs increased as the model complexity increased from YOLOv7-tiny to YOLOv7-X. Among the three original models, the original YOLOv7-tiny model demonstrated the best performance in terms of both accuracy (87.6%) and processing speed (131.6 FPS), which indicated that the Leaky ReLU activation function was more effective than the SiLU activation function, and the smaller anchor size was more suitable for the strawberry dataset. Comparing the model structure of YOLOv7 and YOLOv7-X, the reason for the lower detection accuracy of YOLOv7 was the Extended Efficient Layer Aggregation

Network (ELAN). While ELAN is stable, it is less efficient compared to ELAN-X, which utilizes techniques like expansion, shuffling, and merging cardinality in its computational blocks. The fact that the simplest model, YOLOv7-tiny, delivered the best performance suggests that using overly complex models was unnecessary and would not yield proportional benefits in performance.

*Table 5 The detection performance of YOLOv7, YOLOv7-tiny, and YOLOv7-X*

| Model | FLOPs (G) | mAP$_{0.5}$ F (%) | mAP$_{0.5}$ IF (%) | mAP$_{0.5}$ MF (%) | mAP$_{0.5}$ All (%) | FPS |
|---|---|---|---|---|---|---|
| YOLOv7-tiny | **13.2** | 86.9 | 82.3 | **93.6** | **87.6** | **131.6** |
| YOLOv7 | 105.1 | 77.5 | 70.4 | 86.3 | 78.1 | 47.4 |
| YOLOv7-X | 188.9 | **87.9** | **82.8** | 90.0 | 86.9 | 0.9 |

\* F stands for flower, IF stands for immature fruit, MF stands for mature fruit, and h stands for detection head. Bold means it achieved the best performance in this category.

## 3.3. Performance comparison of Pruning-YOLOv7 series of object detection models

A comparison results of different combinations of detection heads of YOLOv7-tiny, YOLOv7, and YOLOv7-X are shown in Table 6, Table 7, and Table 8, respectively. After reducing the number of detection heads, the network became lightweight, as parts of the layers were useless. Compared with the original model, the best pruning model from each YOLOv7 variants category reveals an average reduction of FLOPs of 21.5%, which indicates that pruning can lighten the model structure and significantly reduce the computation. Pruning-YOLOv7 series of object detection models also have an advancement in both computational efficiency and precision. Compared with the original model, YOLOv7-tiny with head 2 and head 3 had an increase of 10.1% in FPS and an improvement of 1.5% in mAP$_{0.5}$. The fastest model is YOLOv7-tiny with head 3, which achieved 163.9 FPS, and with only a 1.5% decrease in mAP$_{0.5}$. YOLOv7 with detection head 2 improved the $mAP_{0.5}$ to 89.0% and increased FPS to 50.3, compared to the original model's 78.1% $mAP_{0.5}$ and 47.4 FPS, achieving the most significant improvement. The reason can be the scale of detection head 2 of YOLOv7 fitted the best with the size of most objects in the strawberry dataset, while the simplicity of structure made the detection focused on the single detection head range scale and eliminated interference from other detection heads. Compared to YOLOv7-X, the detection accuracy of Pruning-YOLOv7-X with head 2 was 0.5% higher, yet Pruning-YOLOv7-X and its variants exhibited low FPS, rendering them unsuitable for real-time flower and fruit detection.

The detection accuracy of mature fruit consistently surpassed that of immature fruit and flowers. The mature fruit's vivid crimson hue and big size made it distinct from other objects in the strawberry field, however, it may still exhibit a fusion of reddish and whitish due to light reflection, thereby making them look like immature fruits (Fig. 7 (b)). The detection of immature fruit was comparatively challenging due to its similarity to the strawberry leaves, flower, and mature fruits as shown in Fig. 7 (a), and the surrounded by tiny foliage resembling petals. The detection of flowers seemed to be easier than the immature fruit, with the reason that the flower dominated the largest number, and the size mostly focused on medium size. However, small objects are still problems in the detection. Fig. 7 (c) shows that the small size of flowers and immature fruits further exacerbated the challenge in manual labeling and model detection, impeding precise differentiation between flowers and immature fruits.

*Table 6 The detection performance of Pruning-YOLOv7-tiny*

| Model | FLOPs (G) | $mAP_{0.5}$ F (%) | $mAP_{0.5}$ IF (%) | $mAP_{0.5}$ MF (%) | $mAP_{0.5}$ All (%) | FPS |
|---|---|---|---|---|---|---|
| YOLOv7-tiny | 13.2 | 86.9 | 82.3 | **93.6** | 87.6 | 131.6 |
| YOLOv7-tinyh1 | 11.2 | 81.3 | 78.7 | 89.2 | 83.1 | 126.6 |
| YOLOv7-tinyh2 | 10.6 | 89.4 | **83.7** | 93.1 | 88.8 | 128.2 |
| YOLOv7-tinyh3 | **9.9** | 87.1 | 80.6 | 90.6 | 86.1 | **163.9** |
| YOLOv7-tinyh12 | 12.2 | 89.0 | 79.9 | 90.4 | 86.4 | 140.8 |
| YOLOv7-tinyh13 | 12.2 | 89.8 | 82.9 | 91.5 | 88.1 | 135.1 |
| YOLOv7-tinyh23 | 11.6 | **91.6** | 82.8 | 92.8 | **89.1** | 144.9 |

*Table 7 The detection performance of Pruning-YOLOv7*

| Model | FLOPs (G) | $mAP_{0.5}$ F (%) | $mAP_{0.5}$ IF (%) | $mAP_{0.5}$ MF (%) | $mAP_{0.5}$ All (%) | FPS |
|---|---|---|---|---|---|---|
| YOLOv7 | 105.1 | 77.5 | 70.4 | 86.3 | 78.1 | 47.4 |
| YOLOv7h1 | 96.6 | 85.9 | 80.3 | 91.5 | 85.9 | 48.5 |
| YOLOv7h2 | 91.8 | **89.7** | 82.2 | **94.9** | **89.0** | 50.3 |
| YOLOv7h3 | **87.1** | 86.0 | 81.8 | 91.5 | 86.4 | **52.4** |
| YOLOv7h12 | 100.8 | 76.1 | 63.6 | 83.0 | 74.2 | 47.4 |
| YOLOv7h13 | 100.9 | 80.1 | 78.9 | 85.6 | 81.5 | 48.1 |
| YOLOv7h23 | 96.9 | 87.2 | **82.9** | 91.7 | 87.3 | 51.0 |

*Table 8 The detection performance of Pruning-YOLOv7-X*

| Model | FLOPs (G) | $mAP_{0.5}$ F (%) | $mAP_{0.5}$ IF (%) | $mAP_{0.5}$ MF (%) | $mAP_{0.5}$ All (%) | FPS |
|---|---|---|---|---|---|---|
| YOLOv7-X | 188.9 | 87.9 | 82.8 | 90.0 | 86.9 | 0.9 |
| YOLOv7-Xh1 | 176.1 | 86.0 | 81.7 | 94.3 | 87.4 | 0.9 |
| YOLOv7-Xh2 | 161.3 | 88.8 | 81.0 | **92.4** | **87.4** | 1.0 |
| YOLOv7-Xh3 | **146.5** | 88.3 | **84.1** | 88.2 | 86.9 | **1.1** |
| YOLOv7-Xh12 | 182.0 | 74.1 | 60.2 | 82.4 | 72.2 | 1.0 |
| YOLOv7-Xh13 | 182.1 | 81.7 | 71.8 | 83.6 | 79.1 | 1.0 |
| YOLOv7-Xh23 | 167.3 | **89.4** | 79.0 | 87.3 | 85.2 | 1.0 |

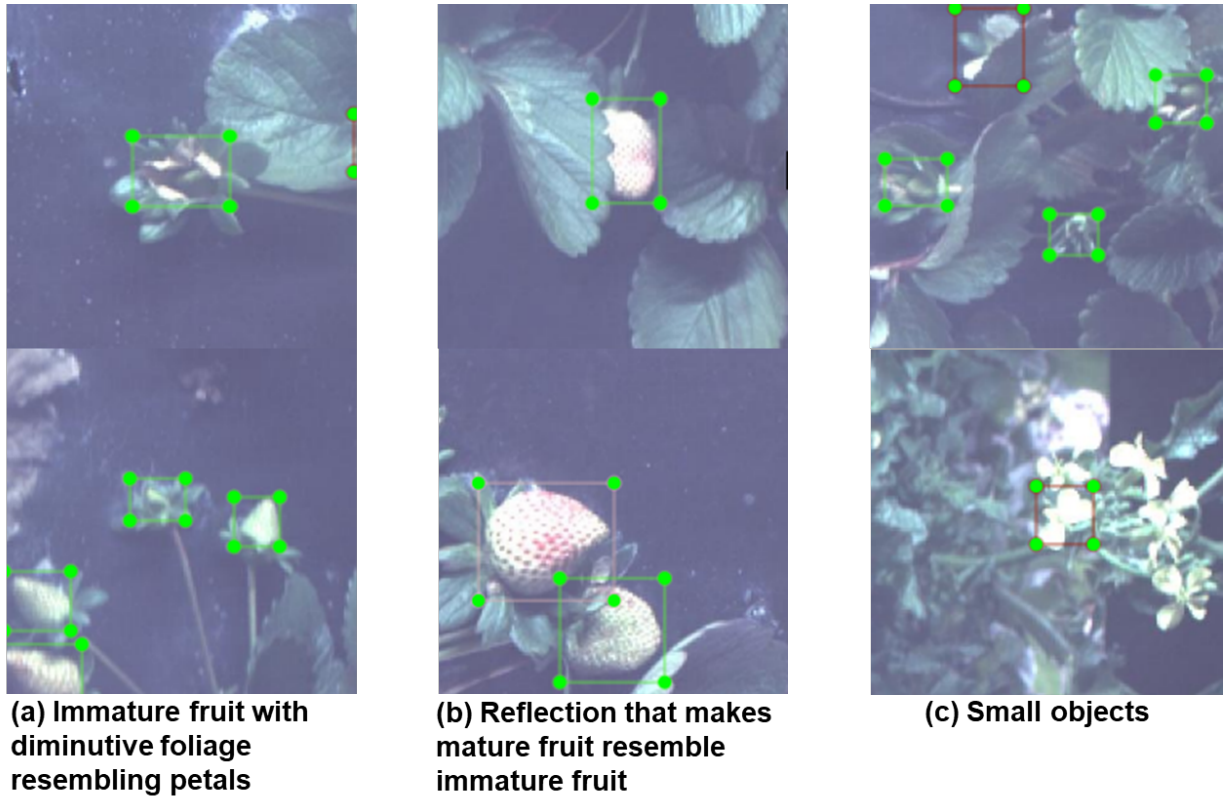

**(a) Immature fruit with diminutive foliage resembling petals**

**(b) Reflection that makes mature fruit resemble immature fruit**

**(c) Small objects**

Fig. 7. Example images showing the three challenges in the detection of immature fruit

To elucidate the rationale behind the enhanced detection performance observed both in inference speed and detection accuracy, the detection outcomes before the Non-Maximum Suppression (NMS) stage of each original model and the corresponding optimal pruning model were compared. Fig. 8 shows that the Pruning-YOLOv7 reduced the number of bounding boxes before the NMS operation. It was observed that a decreased quantity of detection heads led to the reduction of bounding box numbers and an increase in the confidence score, which may hold the potential to streamline the NMS procedure, mitigate potential false positive detection, and reduce the computation. After the NMS operation, the model will generate new images with the final detection result (Fig. 9). The result indicates that Pruning-YOLOv7 can be more adaptable than the original YOLOv7 to scenarios where there are fewer types of target objects. As illustrated in Fig. 9, the detection result of the best model (YOLOv7-tiny with detection heads 2 and 3) was the same as the ground truth, while the other models tended to have a higher incidence of over-detection rather than under-detection.

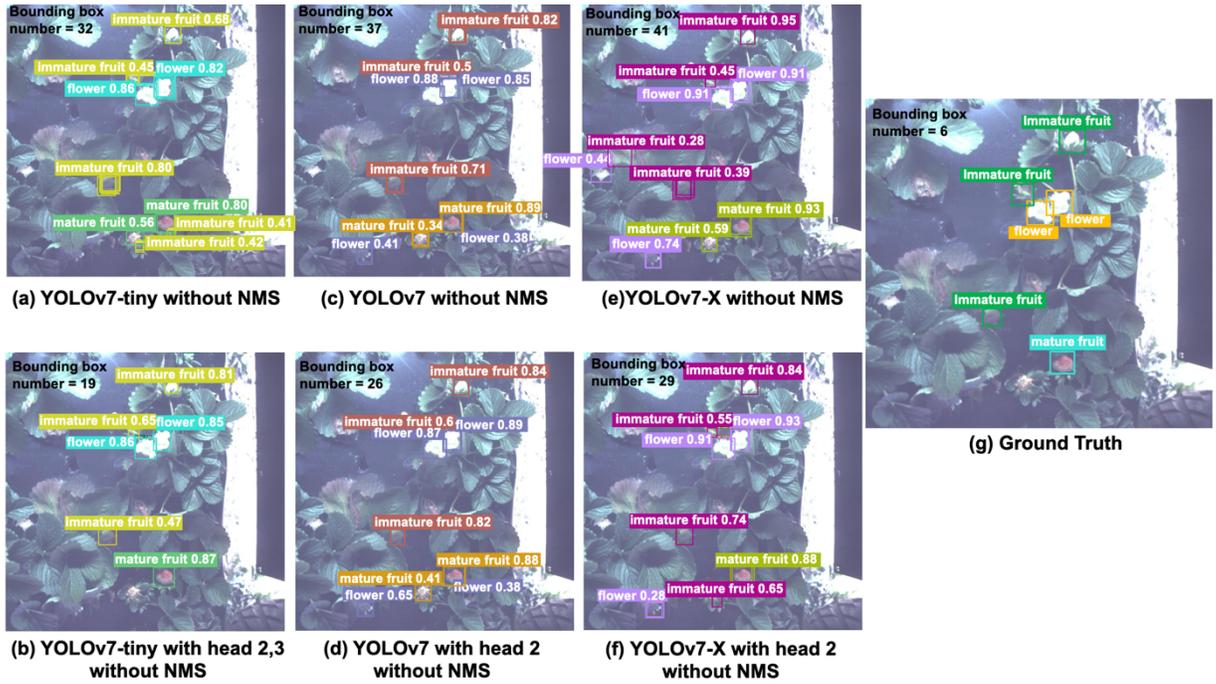

Fig. 8. Pruning-YOLOv7 performance without NMS

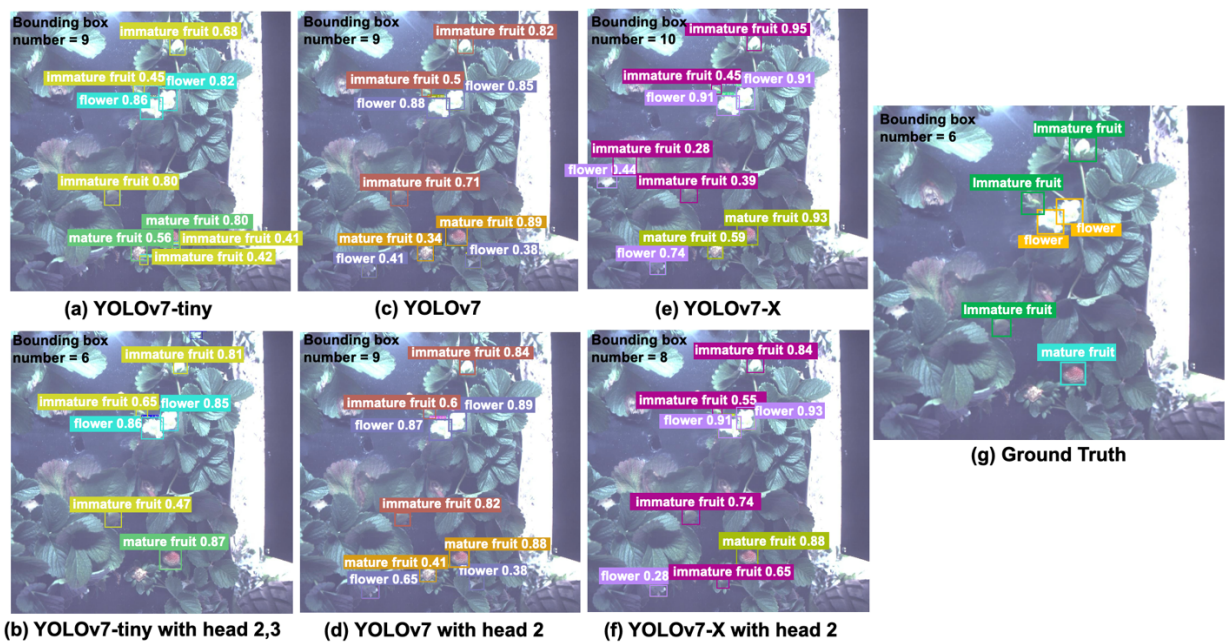

Fig. 9. Pruning-YOLOv7 output

In future studies, the modification of the number of detection heads of YOLOv7 and its variants can be combined with the modification of its backbone part. Lightweight networks, such as EfficientNet ([Tan & Le 2020](#)) can be utilized to replace the backbone of the YOLOv7 to further improve inference speed. Besides, The Pruning-YOLOv7 will be tested on edge devices for real-time application in the strawberry field, and on other crop datasets collected by the on-ground machinery and Unmanned Aerial Vehicles (UAVs), which can further test the applicability of pruning YOLOv7 model heads in different applications.

## 3.4. Flower and fruit tracking

The centroid tracking algorithm (CTA) and IBTA were tested on 55 videos. The details of CTA can be found in previous research ([Puranik et al 2021](#)). The comparison of CTA and IBTA can be found in [Table 8](#). The average MOTA and MOTP of IBTA across all classes were 12.3% and 6.0% higher than that of CTA, respectively. In addition, IDF1, IDR, IDP, and MT of IBTA were all higher than CTA, indicating that IBTA performs better than CTA in identity maintenance and tracking consistency. IBTA also performed better in terms of the number of ID switches, which was only 34 times, compared with 46 times for CTA, showing that IBTA was less likely to mistakenly switch the identities of two targets to each other when tracking objects.

For the mature fruit, the MOTA of both IBTA and CTA was 92.2%, which showed that the two algorithms have the same tracking capabilities in this category. However, IBTA was more accurate than CTA in terms of MOTP. The MOTP of IBTA and CTA were 96.7% and 89.3%, respectively. The tracking of immature fruits and flowers was more challenging as they were numerous and often clustering and the accuracy of the object detection model for them was low compared to mature fruit. Nonetheless, the IBTA was still effective in immature fruits and flower tracking. For immature fruits, the MOTA of IBTA (85.2%) was significantly higher than CTA (58.2%). In terms of MOTP, IBTA was only 2.7% higher than CTA. In the flower category, IBTA's MOTA was 89.5%, nearly 10% higher than CTA's 79.6%. And MOTP of IBTA was 78.4%, which was 7.8% higher than that of CTA (70.6%). For other evaluation metrics, IBTA always achieved the same or even better tracking performance than CTA. This study shows that IBTA consistently outperformed CTA in tracking flower, immature fruit, and mature fruit, respectively.

Inspired by the Deep SORT, IBTA can be improved by incorporating deep learning to extract features and improve tracking performance in future works. Not only on strawberries but IBTA can also be tested on multiple crops and based on other implement platforms and image capture methods. Besides tracking accuracy and precision, tracking speed can also be considered an improvement.

*Table 8 Evaluation of object tracking*

| Model | Class | MOTA↑ | MOTP↑ | IDF1↑ | IDR↑ | IDP↑ | MT↑ | IDs↓ |
|---|---|---|---|---|---|---|---|---|
| | Mature fruit | 92.2% | 96.7% | 90.0% | 90.2% | 89.8% | 22 | 6 |
| IBTA | Immature fruit | 85.2% | 88.2% | 86.2% | 87.2% | 85.2% | 96 | 54 |
| | Flower | 89.5% | 78.4% | 88.4% | 90.2% | 86.7% | 49 | 43 |
| | All | 89.0% | 87.8% | 88.2% | 89.2% | 87.2% | 56 | 34 |
| | Mature fruit | 92.2% | 89.3% | 90.0% | 90.2% | 89.8% | 22 | 6 |
| CTA | Immature fruit | 58.2% | 85.5% | 56.2% | 57.5% | 54.9% | 95 | 76 |
| | Flower | 79.6% | 70.6% | 81.3% | 83.1% | 79.6% | 48 | 56 |
| | All | 76.7% | 81.8% | 75.8% | 76.9% | 74.8% | 55 | 46 |

*↑ indicates the larger the better; ↓ indicates the smaller the better.

## 4. Conclusions

Based on the characteristics of the strawberry datasets, this study developed a simple and effective object detection and tracking method for strawberry yield mapping and prediction. For flower and fruit detection, three Deep learning models (YOLOv7, YOLOv7-tiny, and YOLOv7-X) were first compared, and the result shows that the simplest model, YOLOv7-tiny, achieved the best detection accuracy (87.6%) and fastest inference speed (131.6 FPS). Subsequently, the various combinations of detection heads of deep learning models were evaluated. Notably, YOLOv7-tiny with head 2 and head 3 achieved the best detection accuracy of 89.1% and a high inference speed of 144.9 FPS. This detection result was then used as an input to the flower and fruit tracking method (IBTA), which utilized object spatial information and known speed and direction information of the image acquisition system and achieved better performance than CTA for flower and fruit counting. This study showed that the deep learning model can be optimized by modifying the number of detection heads and combining the known information to make more accurate tracking, and provided a new method to reduce the model complexity without compromising the model performance by utilizing the unique characteristics of the agricultural dataset, which can help monitor the growth and estimate the yield of strawberries in real-time in the future.

## Acknowledgments

The authors would like to thank the Florida Strawberry Research and Education Foundation and the Florida Foundation Seed Producers, Inc. for providing funding support for this study. The authors would also like to thank Ms. Valentina Oropeza for her help in labeling images and videos in this study.

## Appendix

Fig. 10, Fig. 11, Fig. 12, and Fig. 13 show the structure of YOLOv7, YOLOv7-X, Pruning-
YOLOv7, and Pruning-YOLOv7-X, respectively.

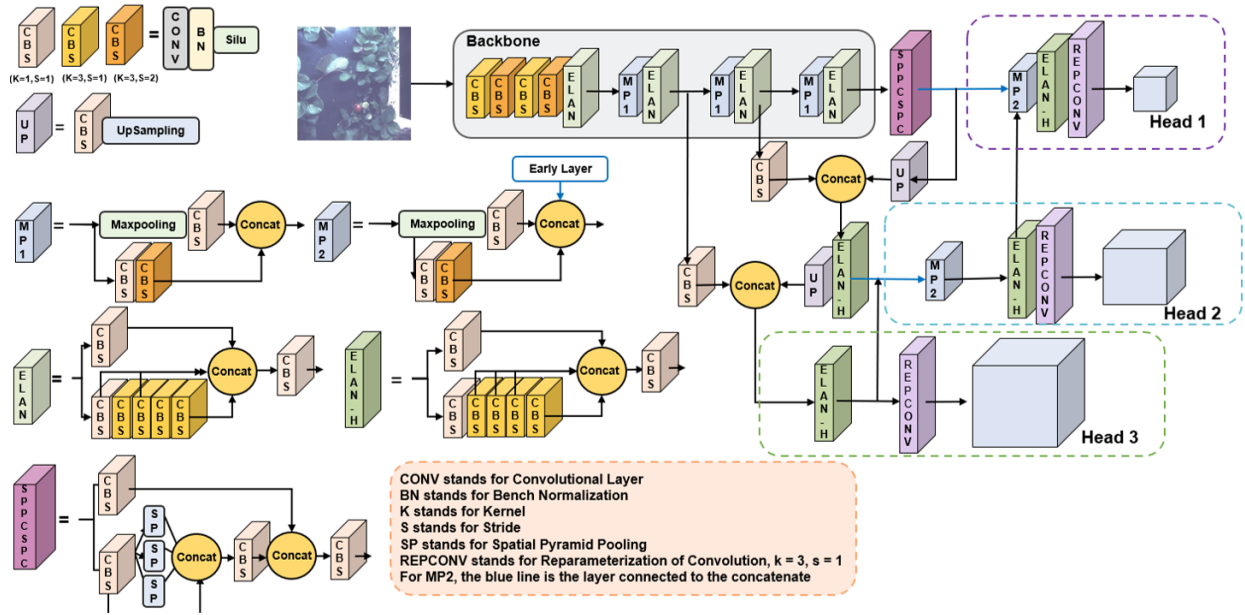

Fig. 10. YOLOv7 structure

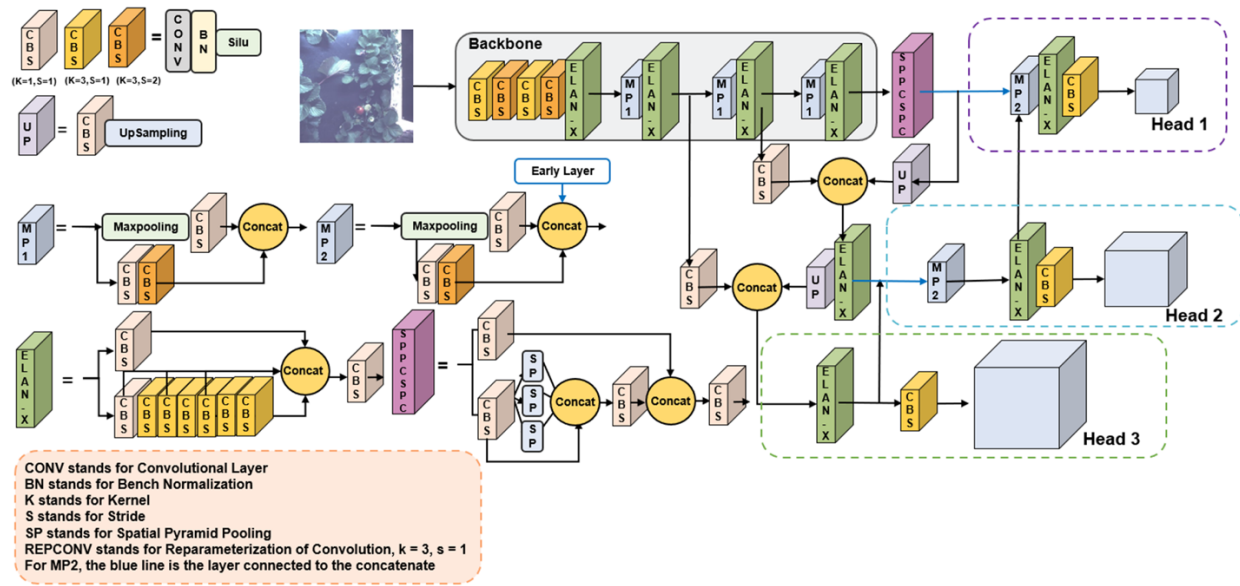

Fig. 11. YOLOv7-X structure

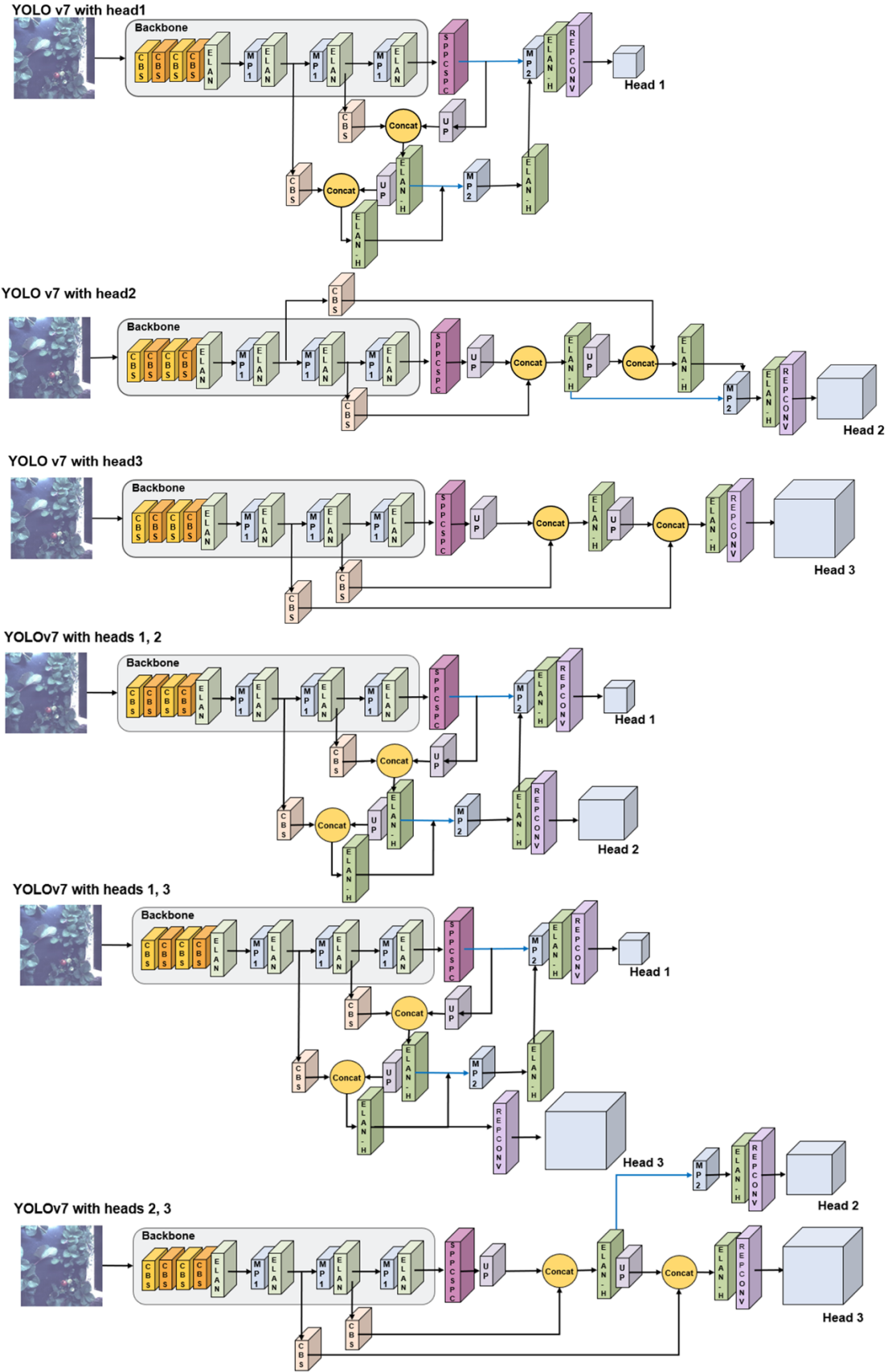

Fig. 12. Pruning-YOLOv7 structure

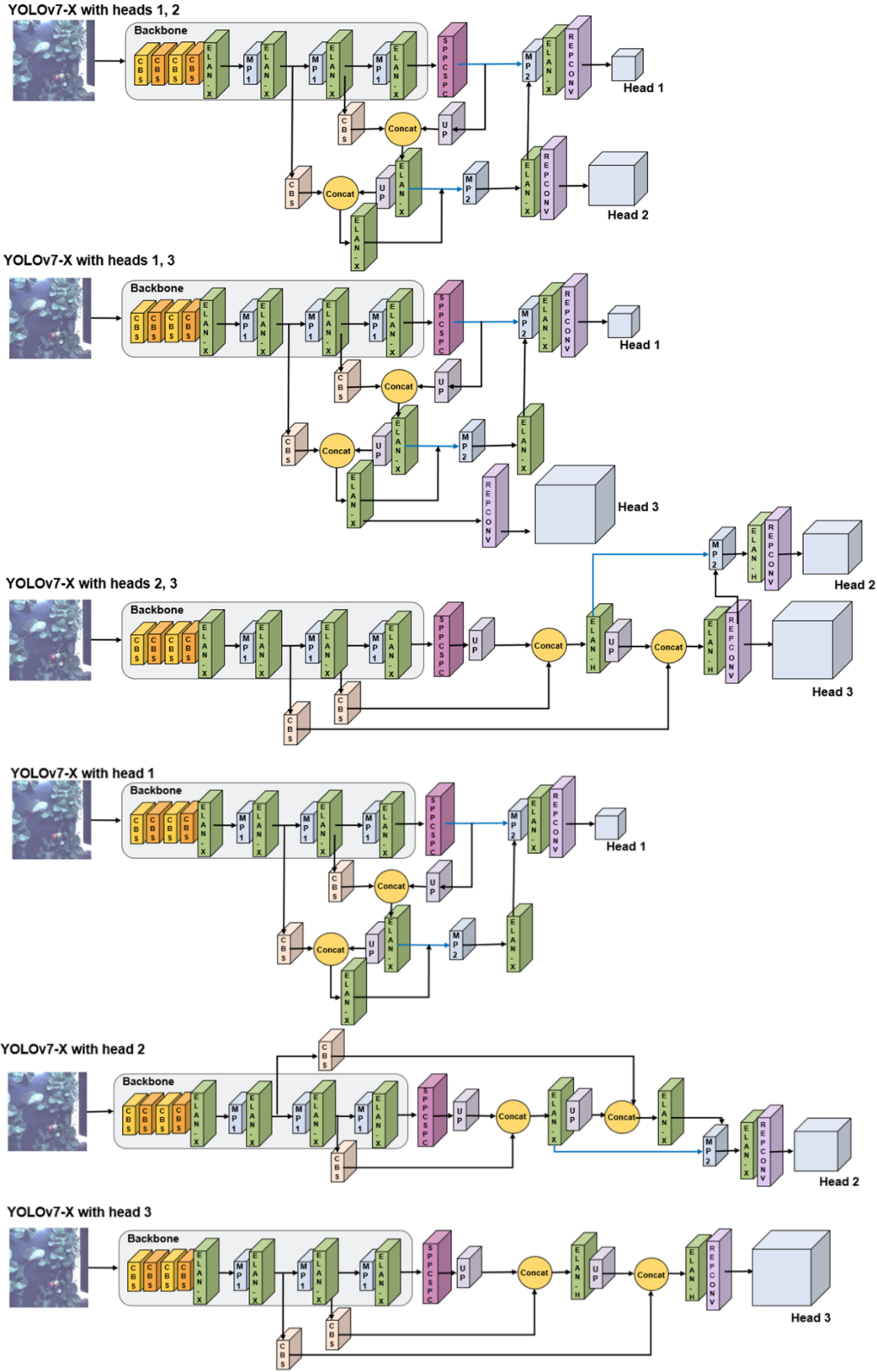

Fig. 13. Pruning-YOLOv7-X structure.